\documentclass[letterpaper, 10 pt, conference]{ieeeconf}  

\IEEEoverridecommandlockouts                              

\usepackage{graphics} 
\usepackage{epsfig} 
\usepackage{times} 
\usepackage{amsmath,amsfonts,amssymb,mathtools,leftidx}  
\usepackage{nccmath} 
\usepackage{mathrsfs}
\usepackage{xcolor}
\usepackage[ruled,vlined,linesnumbered]{algorithm2e}
\usepackage{cite}
\usepackage{url}
\usepackage{todonotes}
\usepackage{booktabs}
\usepackage[a-1b]{pdfx}

\title{\LARGE \bf
Online Modeling and Control of Soft Multi-fingered Grippers \\via Koopman Operator Theory
}

\author{Lu Shi, Caio Mucchiani, and Konstantinos Karydis
\thanks{The authors are with the Dept. of Electrical and Computer Engineering, University of California, Riverside. 
	Email: {\{lshi024, caiocesr, karydis\}@ucr.edu}. 
We gratefully acknowledge the support of NSF under grants \# IIS-1910087 and \# CMMI-2133084. Any opinions, findings, and conclusions or recommendations expressed in this material are those of the authors and do not necessarily reflect the views of the funding agencies.}
}

\begin{document}

\maketitle
\thispagestyle{empty}
\pagestyle{empty}

\begin{abstract}

Soft grippers are gaining momentum across applications due to their flexibility and dexterity. However, the infinite-dimensionality and non-linearity associated with soft robots challenge modeling and closed-loop control of soft grippers to perform grasping tasks. To solve this problem, data-driven methods have been proposed. Most data-driven methods rely on intensive model learning in simulation or offline, and as such it may be hard to generalize across different settings not explicitly trained upon and in physical robot testing where online control is required. In this paper, we propose an online modeling and control algorithm that utilizes Koopman operator theory to update an estimated model of the underlying dynamics at each time step in real-time. The learned and continuously updated models are then embedded into an online Model Predictive Control (MPC) structure and deployed onto soft multi-fingered robotic grippers. To evaluate the performance, the prediction accuracy of our approach is first compared against other model-extraction methods among different datasets. Next, the online modeling and control algorithm is tested experimentally with a soft 3-fingered gripper grasping objects of various shapes and weights unknown to the controller initially. Results indicate high success ratio in grasping different objects using the proposed method. Sample trials can be viewed at \url{https://youtu.be/i2hCMX7zSKQ}.

\end{abstract}

\section{Introduction}
Grasping and manipulation are fundamental functions of robotics. 
Traditional robotic grippers consist of a set of mostly rigid joints and links for a variety of tasks such as perching \cite{backus2014design,liu2020adaptive}, digging \cite{tong2021dig,danfeng2010research}, sorting \cite{gupta2012using,wu2019coordinated}, 
and others \cite{mucchiani2021dynamic,kim2019shallow}. 
Soft grippers offer flexibility and dexterity, and have gained momentum recently (e.g.,~\cite{shintake2018soft,giannaccini2014variable}). Various applications have been increasingly using soft grippers, such as minimally invasive surgery \cite{rateni2015design}, agriculture \cite{navas2021soft}, human rehabilitation \cite{zhang2017design},
as well as manufacturing techniques (3d printing \cite{tawk20193d,wang20163d}, silicone molding \cite{liu2020two, zhou2017soft}, surface and shape deposition \cite{suresh2015surface}), and others \cite{gao2021soft,roels2019multi,glick2018soft}.

Soft robots (including soft grippers) have garnered significant attention and grown into an important research topic, one of the prevailing challenges relates to their modeling and control~\cite{hughes2016soft}. Due to the infinite degrees of freedom of soft robots, accurate first-principles-based models are hard to derive. Moreover, in most cases, system non-linearities pose additional difficulties to control design. One way to resolve such constraints is by considering data-driven methods. 

Various methods based on neural networks have been proposed for modeling and control of soft grippers. The authors in~\cite{zimmer2019ML} show the efficiency of recurrent neural networks in terms of grasp success and stability using a shape-memory soft gripper with integrated hybrid microelectronic sensing skin. Uniformly sampled x-y positions and z axis orientations are considered as candidates in the dataset while no closed-loop control is implemented. In~\cite{gupta2016ML}, a highly compliant multi-fingered hand is controlled by inflating or deflating chambers to demonstrate tasks that human's hand could do, e.g., turning a valve and grasping a bottle from a table. The control algorithm is designed based on imitation learning and reinforcement learning that utilize measurements and experience to update the policy. Another work~\cite{khin2021ML} tries to estimate objects' instability and slippage due to acceleration of the robot or insufficient grasp strength using neural networks to improve grasp performance.

\begin{figure}[!t]
\vspace{6pt}
    \centering
    \includegraphics[width = 0.21\textwidth]{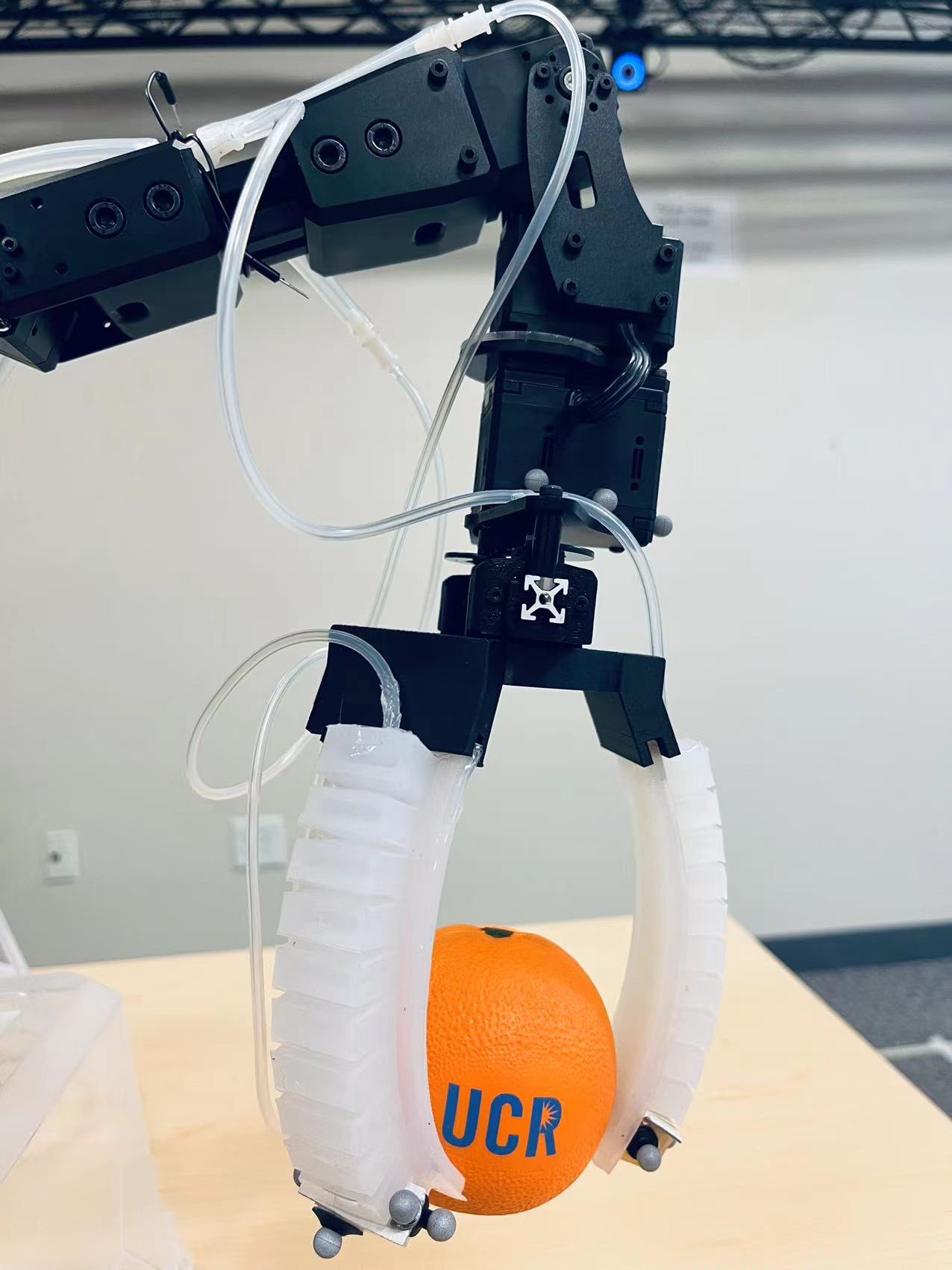}
    \vspace{-8pt}
    \caption{Instance of the soft gripper considered herein grasping a soft ball using the proposed online learning-based method. The method is designed to be agnostic to the shape and/or weight of the object to be grasped.}
    \label{fig:overview}
    \vspace{-18pt}
\end{figure}

An alternative data-driven direction approach, which is also pursued in this paper, is based on the Koopman operator theory. 
The Koopman operator theory can be used to map a complicated nonlinear system into a higher-dimensional linear system approximating the unknown nonlinear dynamics directly from data. Thus, it is appropriate for modeling and control of soft robots~\cite{kaiser2020Overview}. For example, the highly nonlinear dynamics of a soft robot consisting of three pneumatically-driven actuators is identified by Koopman operator~\cite{bruder2019Ksoft}. The method constructs the relationship between PWM signals corresponding to actuator pressure and position/velocity of the end-effector; learned model of this form can then be used directly for control design. In~\cite{haggerty2020Ksoft}, Hankel Dynamic Mode Decomposition is used to estimate the Koopman operator. The learned model is then used in a Linear Quadratic Regulator (LQR) to control a soft robot arm to reach a target position. The Koopman operator theory is also implemented in the modeling of a soft robotic swimmer in simulation~\cite{castano2020Ksoft}.

In this paper, we propose an online modeling and control approach for soft multi-fingered robotic grippers based on Koopman operator theory. First, dictionaries of lifting functions are designed based on the workspace of the soft gripper. Then, instead of implementing the meta-learning based Koopman identifier~\cite{mazouchi2021online}, ACD-EDMD~\cite{shi2021ACD-EDMD} is used to learn and update the model at every time step in real-time without any complicated dictionary learning. Updated models, in turn, provide the dynamics constraints within an MPC structure to help achieve desired control objectives. 

The paper's contribution are twofold:
\begin{itemize}
    \item We propose an online modeling approach that utilizes the Koopman operator theory. An online control algorithm is designed that uses the learned and continuously-refined models in real-time and adapts control inputs for an MPC structure accordingly. 
    \item We evaluate and compare via physical experiments the prediction accuracy of different modeling methods among different datasets using two soft grippers. The proposed online modeling and control algorithm is implemented in the grippers to grasp different objects of varying shape and weight, without any explicit information provided to the algorithm a-priori.
\end{itemize}

\section{Preliminary Material}
\subsection{Koopman Operator Theory and Extended Dynamic Mode Decomposition (EDMD)}\label{sec: EDMD}
The Koopman operator theory~\cite{koopman1931koopman} details how to construct an infinite-dimensional linear operator that governs the evolution of observables $g(\xi_t)$; the latter is also called the ``lifted" variables of original states $\xi_t$.
Then, the original nonlinear system $f$ can be represented by Koopman modes, eigenvalues and eigenfunctions. 
%
%
Consider the nonlinear system $\xi_{t+1} = f(\xi_t)$, 
with $ \xi\in \mathbb{R}^{n_x}$. 
The evolution of observables $g$ using Koopman operator $\mathcal{K}:\mathcal{F} \to \mathcal{F}$ can be written as $\mathcal{K} g(\xi_t)= g(f(\xi_t)).$
Decomposition of the full state observable $g(\xi)=\xi$ using $N$ Koopman modes $v_n$, eigenvalues $\lambda _n$ and eigenfunctions $\varphi_n$, yields~\cite{williams2015EDMD}
\begin{equation}\label{eq:extendPredicion}
    \xi_{t+1} = g(f(\xi_t))= \mathcal{K} g(\xi_t) \to \xi_{t+1}  =\sum_{n=1}^N v_n\lambda_n\varphi_n(\xi_t)\enspace . 
\end{equation}

Given state history {$\Xi = [\xi_1,\xi_2,\dots,\xi_{M},\xi_{M+1}]$} (commonly referred to as ``snapshots"), $\mathcal{K}$ can be expressed as a finite-dimensional approximation $K:\mathcal{F}_N \to \mathcal{F}_N$ of the Koopman operator $\mathcal{K}:\mathcal{F} \to \mathcal{F}$ via EDMD. To do so, we need a dictionary of functions that lift state variables to a higher-dimensional space where observable dynamics is approximately linear. 

%
Given a dictionary $\mathbf{\Psi}(\xi_m)=[\psi_{1}(\xi_m), \ldots, \psi_{N}(\xi_m)]$, the Koopman operator can be approximated by minimizing the total residual between snapshots, 
$J =\frac{1}{2} \sum_{m=1}^{M}\left(\mathbf{\Psi}\left(\xi_{m+1}\right)-\mathbf{\Psi}\left(\xi_{m}\right) K\right)^{2}$. 
This can be solved by truncated singular value decomposition, yielding
    \begin{equation}\label{eq:estimation}
        K \triangleq \boldsymbol{G}^{\dagger} \boldsymbol{A}, \enspace
        \text{where }
        \begin{cases}
      \boldsymbol{G}=\frac{1}{M} \sum_{m=1}^{M}  \mathbf{\Psi}_m^{*} \mathbf{\Psi}_m\enspace,\\
      \boldsymbol{A}=\frac{1}{M} \sum_{m=1}^{M}  \mathbf{\Psi}_m^{*}  \mathbf{\Psi}_{m+1}\enspace,
    \end{cases}
    \end{equation}
with $^\dagger$, $T$ and $^*$ denoting pseudoinverse, transpose and conjugate transpose, respectively. 
With $K$ via~\eqref{eq:estimation}, we obtain 
\begin{equation}\label{eq:KoopmanDecomposition}
    \begin{cases}
        v_n = (w_n^*B)^T \enspace,\\
        \lambda_n \eta_n = K\eta_n \enspace,\\
        \varphi_n = \mathbf{\Psi}_t\eta_n \enspace,
    \end{cases}
\end{equation}
where $\eta_n$ is the $n$-th eigenvector, $w_n$ is the $n$-th left eigenvector of $K$ scaled so $w_n^T\eta_n = 1$, and $B$ is a weight matrix such that $\xi=(\mathbf{\Psi} B)^T$~\cite{williams2015EDMD}. 

The evolution of the original nonlinear system using the estimated Koopman operator is described by replacing expressions~\eqref{eq:KoopmanDecomposition} into~\eqref{eq:extendPredicion}. Control inputs can be readily incorporated to the definition of $\mathbf{\Psi}$ as an augmented state~\cite{proctor2018EDMDc}.

\subsection{ACD-EDMD}
One of the most critical steps that affects the performance of Koopman operator-based methods is how to choose a proper dictionary for lifting the original states, $\mathcal{D} = span \left\{\psi_{1},\psi_{2}, \ldots, \psi_{N}\right\}$. 
In our previous work~\cite{shi2021ACD-EDMD}, we have proposed ACD-EDMD as a general and analytical approach to formalize the construction of the dictionary of lifting functions based on fundamental dynamical system properties, namely the configuration space of a rigid robot and the workspace for a soft robot. 
Hermite polynomials are the basis functions for Euclidean spaces, whereas for non-Euclidean spaces, state variables are mapped into a higher Euclidean space. Then, the dictionary of states $D(\xi)$ is expressed as the Kronecker product of the lifting functions of all these lower-dimensional spaces. For the control input, $D\mathbf{u})$ is computed by the zero- and first-order Hermite polynomials and Kronecker products thereof similarly to the $\mathbb{R}^{n_x}$ cases for states. Finally, the complete dictionary is formed as the Kronecker product between lifting functions for states and inputs, i.e. $\mathcal{D} = kron(D(\xi), D(\mathbf{u}))$. 
With the proper design of the lifting functions, we can then apply the algorithm introduced before in Section~\ref{sec: EDMD} to extract a model of the robotic system at hand using the Koopman operator theory. 

\section{Soft Gripper System Design and Properties}

In this work we consider soft multi-fingered robotic grippers that are pneumatically actuated. A rigid sub-system is used for gross transport and positioning of the wrist plate, which contains the soft fingers (actuators) that are responsible for object grasping (Fig.~\ref{fig:overview}). Note that motion of the rigid sub-system is not included in the model we aim to learn. Instead, we only consider the grasping behavior of the soft fingers in this paper. 


\subsection{Soft Gripper Design and Integration}
\begin{figure}[t]
    \vspace{6pt}
    \centering
    \includegraphics[width = 0.305\textwidth]{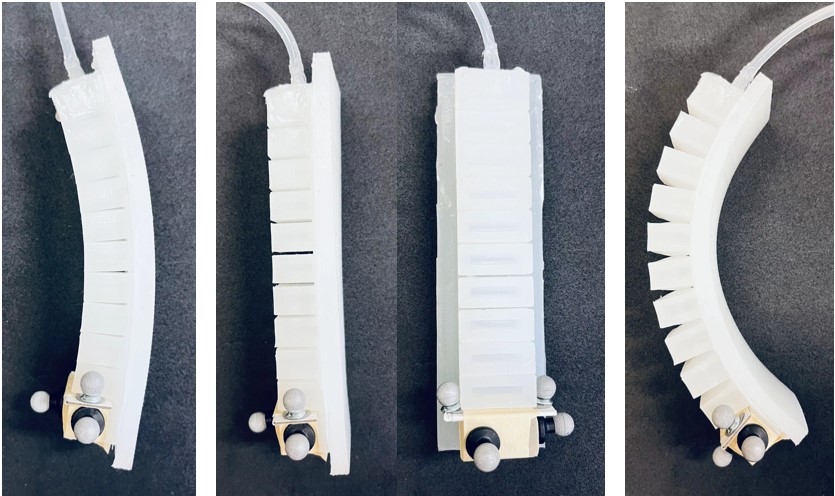}
    \vspace{-6pt}
    \caption{Soft finger depressurized (left), idle (middle) and pressurized (right).}\label{fig:actuator}
    \vspace{-16pt}
\end{figure}

\begin{figure}[h]
\vspace{-6pt}
    \centering
    \includegraphics[width = 0.14\textwidth]{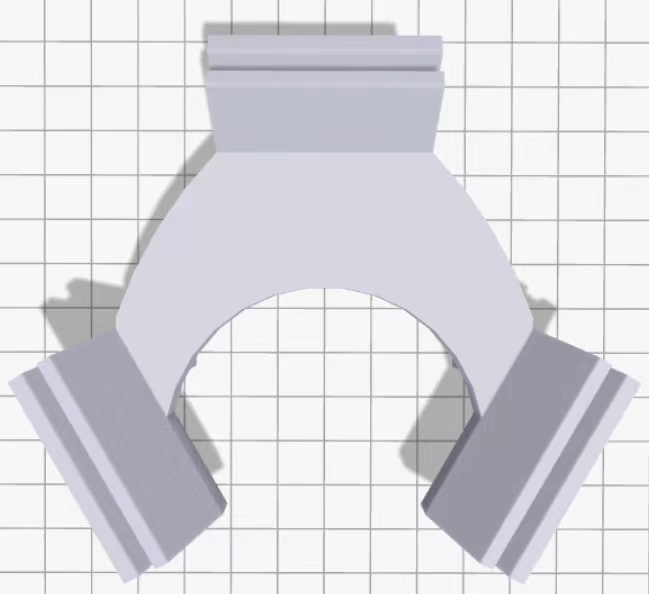} \hspace{5pt}
    \includegraphics[width = 0.156\textwidth]{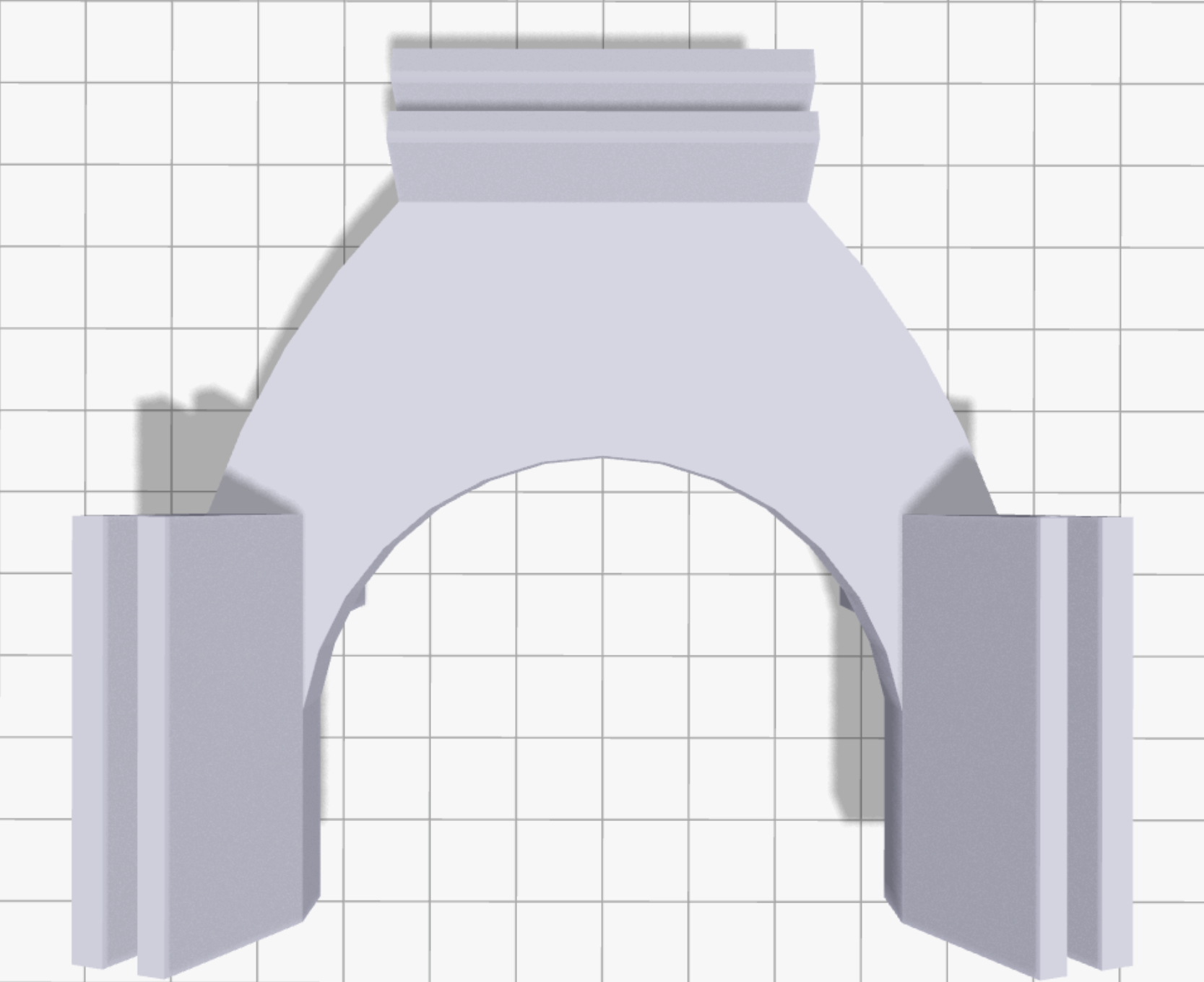}
    \vspace{-6pt}
    \caption{CAD renderings of the rigid attachment plates considered herein that afford symmetric (left) and asymmetric (right) soft finger arrangements.}
    \label{fig:holders}
    \vspace{-12pt}
\end{figure}

The soft gripper comprises three identical bending actuators~\cite{kokkoni2020finger} (Fig.~\ref{fig:actuator}) which are retrofitted PneuNets~\cite{shepherd2011multigait}. These actuators use pneumatic pressurization and depressurization applied to in-series connected deformable chambers to create rotational motion. The soft actuators are made via silicone (Smooth-On Dragon Skin 30) casting in 3D-printed molds (with carbon-fiber reinforced Onyx filament), and are attached to 3D-printed plates. We consider two distinct plates for arranging the soft fingers, one where the three fingers are arranged in a symmetric manner and one that is asymmetric (Fig.~\ref{fig:holders}). 
The soft gripper assembly is attached to the ReactorX 150, a rigid manipulator with four links and four joints.\footnote{~The original fifth joint of the robot is a rigid pincher which is removed for the purposes of this work.} The plate acts as a wrist affixed to the manipulators endpoint.

The rigid sub-system serves a dual purpose: first to move the soft gripper to a desired position and then to lift the soft gripper after it has grasped an object. Note that the focus of this work is on the soft gripper, assessing how our proposed Koopman operator-based algorithm can enable the soft gripper to grasp objects of different shapes and weight without explicit knowledge of. To evaluate grasping success, we quantify if the gripper can hold the object as it is being lifted up by the rigid manipulator.

\subsection{Pneumatic Actuation}
\begin{figure}[!t]
    \vspace{6pt}
    \centering
    \includegraphics[trim={0cm 4cm 0cm 1cm},clip,width = 0.20\textwidth]{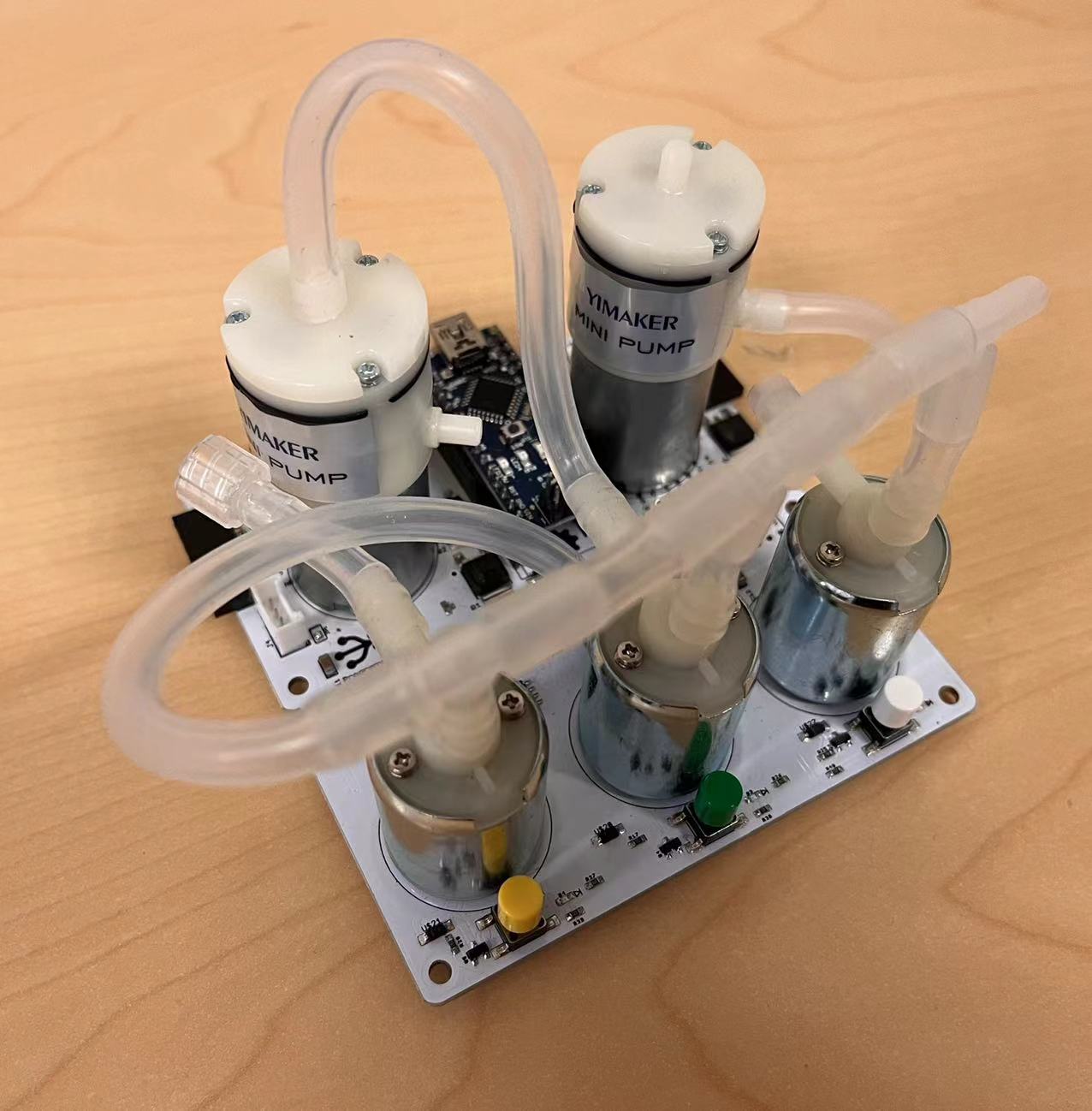}
    \vspace{-6pt}
    \caption{The Programmable-Air Pneumatic Control Board.}
    \label{fig:board}
    \vspace{-18pt}
\end{figure}

The soft fingers are pneumatically actuated using Programmable-Air boards (Fig.~\ref{fig:board}). 
The board is based on the Arduino nano (ATMega328P), and has a maximum pressure of $\pm \; 50 \; kPa$ and flow rate of $2\; liters/min$. Pressurization is controlled by changing the PWM signal sent to pumps. Two such boards were considered for the experiments; the detailed configuration is discussed in Section~\ref{sec:experiment}.

\subsection{Workspace Construction}\label{alg: ACD}
As discussed in~\cite{shi2021ACD-EDMD}, the workspace of soft robots needs to be defined to obtain the dictionary of lifting functions, which is used in the Koopman operator-based modeling approach. 
The workspace of a robot is the set of positions that it can reach. 
For soft multi-fingered grippers, the position of the tips of the fingers can be considered to construct the workspace. The state of the soft gripper is thus defined by the concatenation of the positions of each tip as $\xi_1 = [x_1,y_1,z_1]$, $\xi_2 = [x_2,y_2,z_2]$, and $\xi_3 = [x_3,y_3,z_3]$, expressed in terms of an inertial frame set at the center of the plate (Fig.~\ref{fig:coordinates}).\footnote{~Although this work uses a $3$-fingered soft gripper, the proposed approach can be directly extended to account for general $n$-fingered soft grippers.}

We can further simplify the workspace used for operator learning by exploiting the physical constraints underlying the $\{x_i,y_i,z_i\},i=1,2,3$ state vectors. The types of actuators considered herein are constrained to create planar motion and can only bend about the normal vector to the plane they are constrained to create motion at (which is determined herein based on the arrangement of the rigid attachment plate). This way, only one of $\{x_i,y_i,z_i\}$ for each finger suffices, thus giving rise to a workspace for the whole soft gripper of $\mathbb{R}^1\times \mathbb{R}^1\times \mathbb{R}^1$. Without loss of generality, here we choose $z_i$ as the representations of sub-workspace for each of the fingers, but any other state component would work as well. 
Besides the Kronecker product of sub-dictionaries, we also include in the library observations of other states and inputs for full-observability. We define $D_{\xi} = [1,z
_1,z_2,z_3,z_1 z_2,z_1z_3,z_2z_3,z_1z_2z_3]$. The dictionary is computed as $D = [\xi_1,\xi_2,\xi_3, kron(\mathbf{u},D_{\xi})]$. After eliminating redundant terms, the final dictionary contains $29$ terms.\footnote{~It is beneficial to seek to construct the smallest possible dictionary that best describes the system at hand for computational expediency. At times, doing so relies on exploiting engineering intuition and/or any system symmetries. In this work we take advantage of the bending-only motion afforded by the specific fingers we consider. However, our method is general and can directly apply to any type of soft fingers that afford unconstrained position (and orientation) control of their tips in full 3D space, albeit at the expense of increasing the size of the constructed dictionary.}


\section{Online Modeling and Control}

Unlike rigid robotic systems whereby it is known how to construct their dynamics, soft robotic systems possess complicated and, most often, infinite-dimensional dynamics. Thus, data-driven approaches may be a better fit for extracting models for soft robotic systems directly from data. 
However, models learned offline or in simulation (a technique often employed in relevant works) may be invalidated at runtime due to the presence of uncertainty~\cite{karydis2015IJRR}, unless a very delicately designed experimental environment and operation process are imposed for the dynamics learned offline to remain valid for online model-based control. This process can get very tedious and is not scalable. 
Instead, online control should require limited data and limited time to enable online computation, which are key properties enabled via Koopman operator-based methods~\cite{mazouchi2021online,shi2020CCTA}. 
Thus, we propose an online modeling approach using Koopman operator theory to update dynamics constraints in real time and adjust inputs accordingly within an MPC structure.\footnote{~While we consider MPC in this work, the continuous refined learned model can be used in any other online model-based control algorithm.}

\subsection{Online Model Predictive Control (MPC) Formulation}
MPC determines control inputs at each time step by solving an optimization problem among a range of prediction horizon $H_p$ constrained by robot dynamics. The general optimization algorithm can be formulated as 
\begin{equation*}\label{eq:mpc}
    \min\limits_{u_i} \enspace J((u_i)_{i = 0}^{H_p-1},(\xi_i)_{i = 0}^{H_p},),  
\end{equation*} 
\begin{equation*}
    \begin{cases}
       \xi_{i+1} = f(\xi_i, u_i),\enspace i = 0, \dots,H_p-1, \\
       \xi_0 = \xi_{initial}.
    \end{cases}
\end{equation*}

In this work, the soft grippers' dynamic constraints are obtained via ACD-EDMD in real time. In other words, at each time step, we solve an optimization problem using an updated learned model. The cost function $J$ is usually user-designed. Herein, we set a desired trajectory $r$ and minimize the difference between predicted states and the desired points. With these considerations in place, the MPC problem solved here at each time step $t$ is expressed as 
\begin{equation}\label{eq:acd-mpc}
    \min\limits_{u_i} \enspace J=((u_i)_{i = 0}^{H_p-1},(\xi_i-r_i)_{i = 0}^{H_p},),
\end{equation}
\begin{equation*}
    \begin{cases}
       \xi_{i+1} = \text{ACD-EDMD}(\xi_i, u_i),\enspace i = 0, \dots,H_p-1, \\
       \xi_0 = \xi_t.
    \end{cases}
\end{equation*}

\subsection{Online Modeling and Control Algorithm}
We are now ready to present the proposed online modeling and control algorithm (Alg.~\ref{alg:alg}). After initialization, at each time step $t$, control signals are obtained from the MPC controller with the updated model constraints learned based on data from the previous time step. Then, the physical robot is driven by the control inputs and observations are collected to the online training dataset. A refined model estimated from the updated dataset via ACD-EDMD is then used to create updated dynamic constraints for the following step. The process is repeated until the control task (here grasping different objects) is achieved.

\section{Experimental Testing and Results}\label{sec:experiment}
To test the efficiency of our online modeling and control algorithm, we evaluate separately modeling accuracy and control performance. 
First, we compare the performance of different model-estimation approaches in online and offline datasets to show the benefits of training with online observations. We further compare the ACD-EDMD based online modeling estimation accuracy with SINDy (Sparse Identification of Nonlinear Dynamical Systems (SINDy)~\cite{brunton2016SINDy}) and LSTM (Long Short-Term Memory~\cite{hochreiter1997long}) using online training datasets. 
Then, we implement our online control algorithm in the soft gripper and evaluate it in grasping different objects.

\begin{figure}[!t]
\vspace{6pt}
    \centering
    \includegraphics[trim={0cm 2cm 0cm 0.75cm},clip,width = 0.22\textwidth]{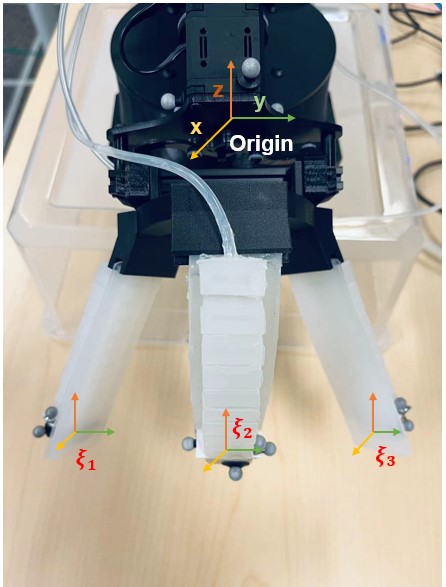}
    \vspace{-6pt}
    \caption{Coordinate systems and motion capture marker configuration.}
    \label{fig:coordinates}
    \vspace{-9pt}
\end{figure}

\newlength{\textfloatsepsave} \setlength{\textfloatsepsave}{\textfloatsep} 

\begin{algorithm}[h]
\caption{Online Modeling and Control}\label{alg:alg}
\textbf{initialize:} Evolve the system with random valid inputs for the first $M$ time steps. Then, compute the estimated Koopman operator from the observations: 
$K = \text{ACD-EDMD}(\{\xi\}_{1}^{M+1},\{u\}_{1}^M)$
\\
\For{$t\geq M$}{
\Repeat{\text{control task is finished}}{
\textit{\textbf{MPC Controller}}:  
Calculate the control inputs with the model constraints learned in the previous step by solving optimization~\eqref{eq:acd-mpc}.
\\
\textit{\textbf{Plant}}: Propagate the soft robotic gripper with $u = [u_1,u_2]$:
    $\xi_{t+1} = f(\xi_{t},u)$
    , where $f$ is the evolving law of real plant.
\\
\textit{\textbf{Model}}:
Collect new data from observations and update the ACD-EDMD model.
\\
\textbf{$t \gets t+1$}
}
}
\end{algorithm}
\setlength{\textfloatsep}{0pt}


\begin{table*}[ht]
\vspace{6pt}
    \centering
    \vspace{-2pt}
    \caption{Prediction accuracy of different approaches}\label{tab:prediction_results}
    \vspace{-7pt}
    \begin{tabular}{p{0.1\textwidth}p{0.125\textwidth}p{0.175\textwidth}p{0.17\textwidth}p{0.17\textwidth}p{0.08\textwidth}}
        \toprule
                    \emph{Method} & \emph{Dataset} &
        MSE($|\xi_1|$) & MSE($|\xi_2|$) & MSE($|\xi_3|$) & (magnitude)\\ 
        \midrule
        SINDy & Offline (Unloaded)& $[0.00,556.66,0.062]$&$[340.17,0.44,188.47]$& $[12.84,327.78,0.014]$& $\times 10^3$\\
        \midrule
        ACD-EDMD & Offline (Unloaded) & $[0.26,256.62,5.76]$& $[180.83,7.12,133.35]$& $[7.91,141.33,0.69]$& $\times 10^2$   \\
        \midrule
        SINDy & Offline (Loaded) &$[364.17,38.57,0.74]$& $[0.95, 0.99, 26.59]$&$ [122.65,0.13,3.25]$&$\times 10^1$ \\
        \midrule
        ACD-EDMD & Offline (Loaded) & $[211.90,75.52,0.0026]$& $[0.43,0.13,0.47]$& $[62.36,15.21,0.05]$& $\times 10^0$ \\
        \midrule
        \midrule
        SINDy & Online & $[0.0047,0.0113,0.0006]$ &$[1.36,1.24,0.09]$ &$ [0.44,0.0022,0.03]$ &$\times 10^0$\\
        \midrule
        LSTM & Online & $[0.58,1.40,0.0136]$&$ [0.38,0.76,0.60]$&$[1.45,1.81,0.87]$& $\times 10^{-2}$ \\
        \midrule
        \textbf{ACD-EDMD} & Online & $[0.32,2.16,1.58]$ &$[0.17, 0.93,0.20]$&$[0.40,1.35,0.15]$&$\times {\bf 10^{-4}}$ \\
        \bottomrule
    \end{tabular}
    \vspace{-18pt}
\end{table*}

Motion capture system (12-camera Prime13 Optitrack) is used to collect position data ($x$, $y$, $z$) of the tip of each finger. To ensure consistency among all training data, we set the origin of the inertial frame at the geometric center of the attachment plate (Fig.~\ref{fig:coordinates}). 
%
%
In both of the grippers, one control board controls soft fingers $1$ and $3$, and the other control board controls finger $2$.  Thus, together with the PWM signals of two control boards, $9$ states and $2$ inputs of the robotic soft gripper are observed. 

\subsection{Settings of Different Approaches to Compare Against}

\subsubsection{SINDy}
SINDy is a data-driven approach to extract governing equations of nonlinear systems. It builds on the assumption that only a few key nonlinear candidates governing the evolution of original system. Sparse regression determines weights of those nonlinear terms from observation data. Here we adopt the example in~\cite{brunton2016SINDy} as the library $\Theta(X)$ of candidate nonlinear functions, consisting of constant, polynomial and trigonometric terms, that is 
\begin{equation*}\label{eq:SINDy fcn}
    \Theta(X) = \left[ \begin{matrix}
    1 & X & X^2 & \sin(X) & \cos(X)
    \end{matrix}\right].
\end{equation*}
The sparse regression problem is solved by least absolute shrinkage and selection operator (LASSO)~\cite{tibshirani1996LASSO}.  

\subsubsection{LSTM} LSTM is an artificial recurrent neural network that offers feedback connections and the ability to deal with entire datasets rather than single data points. 
LSTM networks have been used in robotics applications (e.g.,~\cite{park2018multimodal}),  notably in-hand robotic manipulation~\cite{andrychowicz2020learning}. In this paper, we use the LSTM layer in MATLAB R2020a Deep Learning Toolbox to construct the network. A single layer of 20 hidden units is applied as designed in~\cite{zimmer2019ML}.

\subsection{Model Estimation Accuracy}
Different datasets for training are collected as follows.

\begin{figure}[!t]
    \vspace{4pt}
    \centering
    \includegraphics[width = 0.47\textwidth]{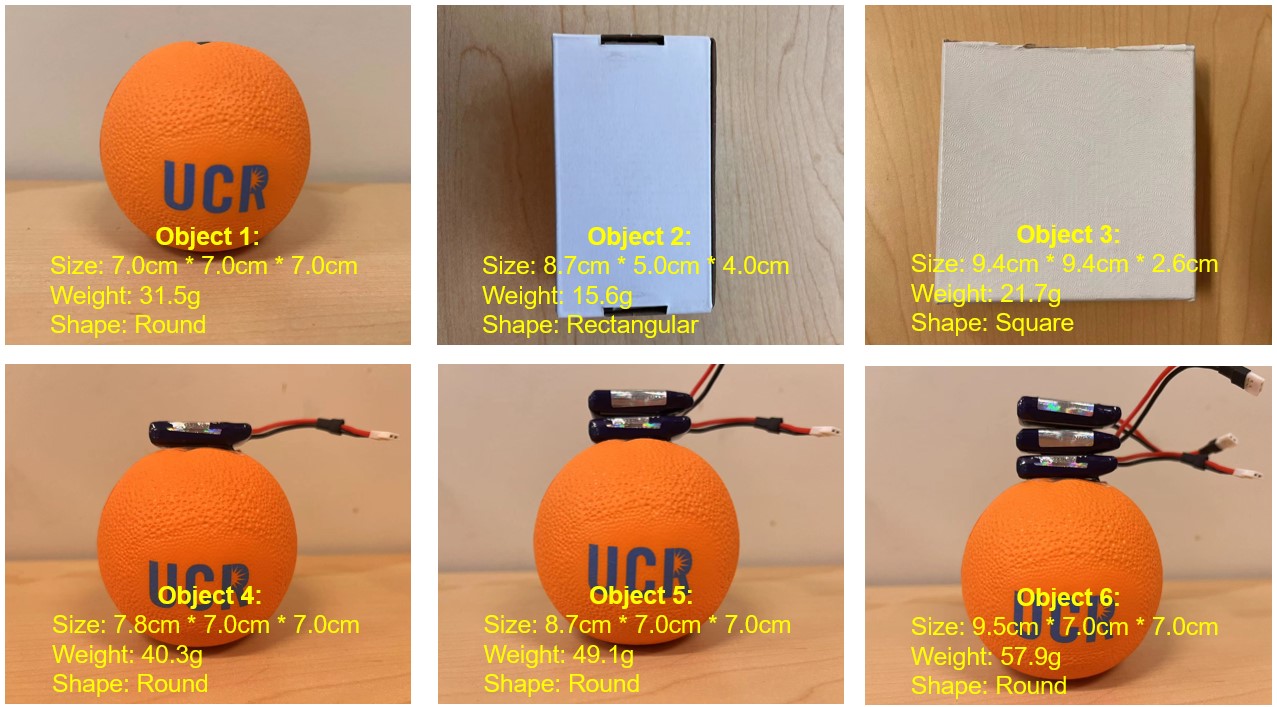}
    \vspace{-9pt}
    \caption{Objects of different shape and/or weight considered in this work. Sample trials can be viewed at \url{https://youtu.be/i2hCMX7zSKQ}.}
    \label{fig:objects}
    \vspace{-12pt}
\end{figure}

\begin{enumerate}
    \item \textbf{Unloaded Offline training data}: In offline training, constant PWM signals are given by the control board. Each signal lasts for $16$ sec with a sampling rate of $2$ Hz ($0.5$ sec). Preliminary testing revealed that the finger can function effectively when the PWM is within the range of $20\%$ to $35\%$. Thus, we collect the position and voltage signals data by either pressurizing or depressurizing all the fingers when the gripper is unloaded within that range. A total number of $32$ sets are collected as the `Unloaded Offline' training data. A moving-average filter with window size of $5$ is applied to the offline training data.
    
    \item \textbf{Loaded Offline training data}: Data from four trials of grabbing plush oranges (object $1$ in Fig.~\ref{fig:objects}), two trials of grabbing the smaller rectangular box (object $2$ in Fig.~\ref{fig:objects}) and two trials of grabbing the larger square box (object $3$ in Fig.~\ref{fig:objects}) are collected with length of $26$ points for each trial. The `Loaded Offline data' case is constructed as a combination of these $8$ data sets.
    
    \item \textbf{Online training data}: Online observations are collected when the gripper attempts to grasp the plush orange. Let $N_T$ be the training data length; every $\xi_{t-N_T-1}^{t}$ observations are used to predict $\xi_{t+1}$.
\end{enumerate}

The three aforementioned datasets are used to predict $\xi_{t+1}$ for the whole online data trial of length $L$.
We calculate the Mean Squared Error (MSE) between predicted (superscript $p$) and true (superscript $t$) states per $MSE = \frac{1}{L-N_T}\sum_{N_T}^L ([\xi_1^{p}, \xi_2^{p},\xi_3^{p}]-[\xi_1^t,\xi_2^{t},\xi_3^{t}])^2$ for the whole trajectory. Results are contained in Table~\ref{tab:prediction_results}. For all algorithms, predictions computed from the offline \textit{loaded} data are more accurate than those made based on \textit{unloaded} data. Further, \textbf{online} approaches lead to significantly smaller error than all \textbf{offline} approaches.

We also compare the efficiency of all methods with online training data by evaluating repeated grasping of object $1$ for $8$ times to avoid bias. We tested all three approaches with different training data size, $N_T = \{3,5,10\}$. 

\begin{figure*}[ht]
\vspace{6pt}
    \centering
 \includegraphics[width = 0.85\textwidth]{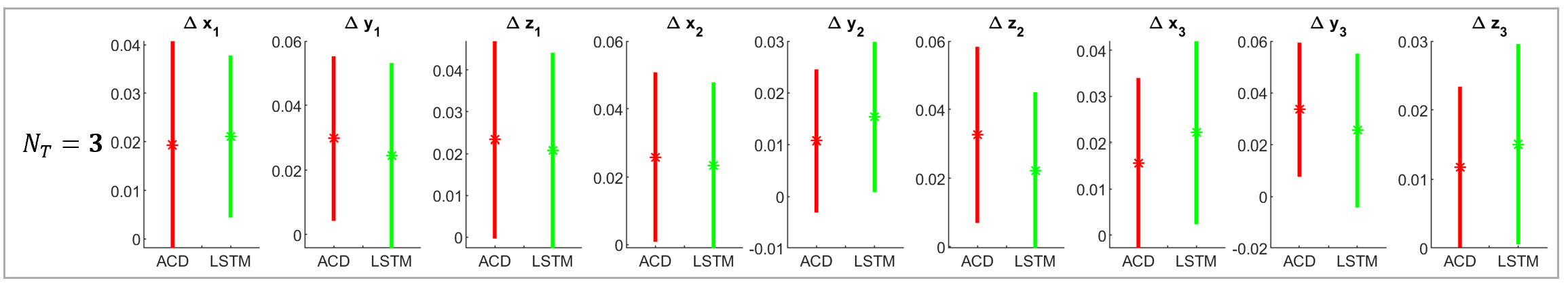}
  \includegraphics[width = 0.85\textwidth]{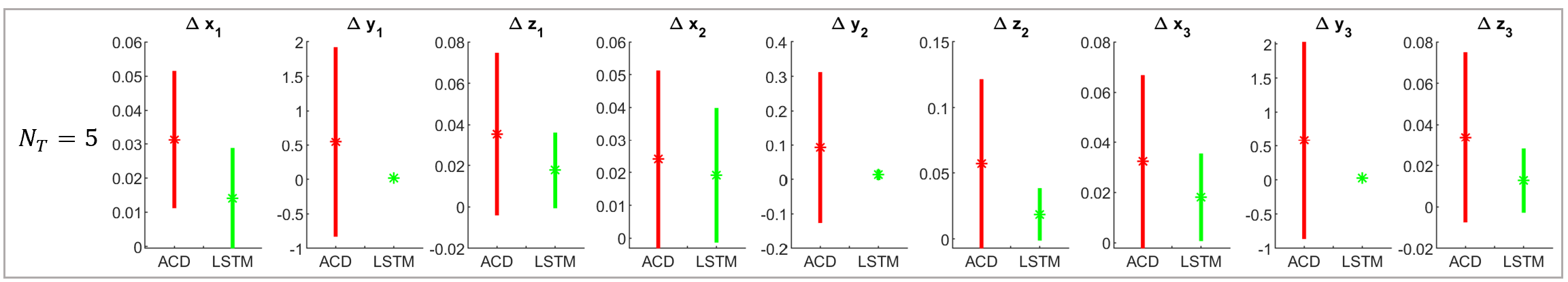}
  \includegraphics[width = 0.85\textwidth]{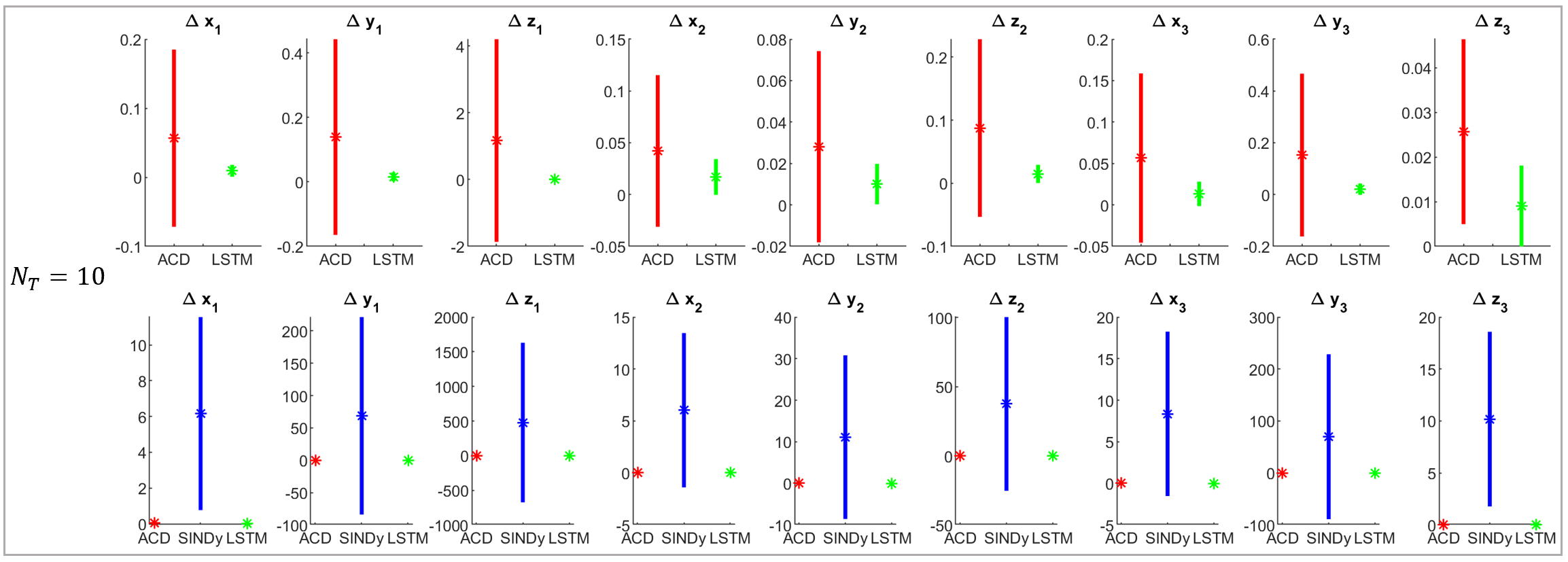}
  \vspace{-9pt}
    \caption{$1$-$\sigma$ error prediction accuracy of all online methods. SINDy (blue) has large errors; to keep the bars visible, one case ($N_T=10$) is shown.}
    \label{fig:Results}
    \vspace{-18pt}
\end{figure*}

Results are depicted in Table~\ref{tab:time} (training time) and Fig.~\ref{fig:Results} (prediction accuracy). 
Our algorithm can run $2\times$ and $3\times$ {\bf orders of magnitude faster} than SINDy and LSTM, respectively, per each time step. This performance is attainable because ACD-EDMD leverages fundamental characteristics of the dynamical system at hand (herein the workspace of the soft gripper) which leads to a simultaneously smaller and more descriptive dictionary to use. 
Further, with reference to Fig.~\ref{fig:Results}, our method can achieve similar accuracy with LSTM for smaller training data size, and comparable accuracy for larger training data size. (More specifically, our method can lead to more variable accuracy compared to LSTM for $N_T=10$, but mean values are close.) The combination of comparable accuracy but substantially faster training time makes our proposed approach appropriate for online learning-based modeling and control tasks. 


\setlength{\textfloatsep}{\textfloatsepsave}

\begin{table}[h]
\vspace{-6pt}
    \centering
    \caption{Training times per each time step for different approaches}\label{tab:time}
    \vspace{-6pt}
    \begin{tabular}{c c c c}
    \toprule
     Training Size \textbackslash~ Approach&  ACD-EDMD & SINDy & LSTM\\
     \midrule
    $N_T = 3$ & $0.00028$ s& $0.0260$ s& $0.7240$ s\\
    \midrule
    $N_T = 5$& $0.00030$ s&$0.0583$ s&$0.7837$ s\\
    \midrule
    $N_T = 10$& $0.00044$ s&$0.1542$ s&$0.8468$ s\\
    \bottomrule
\end{tabular}
    \vspace{-6pt}
\end{table}

The bottom row panels of Fig.~\ref{fig:Results} show that the prediction error obtained by SINDy is much larger than the other two approaches.\footnote{~For error bars to be visible, we omitted those of SINDy in $N_T=\{3,5\}$.} 
SINDy's worse than expected performance may be because of the small training data size used in the online procedure, or perhaps the choice of candidate nonlinear functions selected based on the original publication~\cite{brunton2016SINDy}. This is in fact one key point of departure from SINDy in that it requires selection of candidate nonlinear functions and that selection can affect the overall performance.

\subsection{Online Modeling and Control: Grasping}
For the final evaluation of our method, we apply Alg.~\ref{alg:alg} to grasp objects using the two soft grippers with different finger configurations. The online training size $N_T$ is $5$ and the prediction horizon $H_p$ is set at $3$. In this experiment, we manually designed desired trajectories for each finger.\footnote{~Note that this experiment is only one example to show how our online modeling and control approach can be implemented in practice. Other model-based controllers and desired trajectories can be used as well, and one can embed our online modeling method into their design directly, which is one key benefit of our proposed method.} The same desired trajectory is used for grasping objects of different shape and same shape but different weight (Fig.~\ref{fig:objects}), without any information about the objects. Each grasping task is repeated $10$ times to avoid bias. Successful grasps are considered those that the gripper grasps the object, holds it until fully lifted and keeps it for $3$ sec after lifting.
%

Results show that the asymmetric gripper functions better than the symmetric one in terms of grasping heavier objects, whereas the symmetric gripper performs better on grasping objects of varying shape (Table~\ref{tab:success_rate}). 
However, we can still deduce that our algorithm can be easily implemented into different robots to grasp different objects without requiring any knowledge about the robot and the objects.

\section{Conclusions}
Soft multi-fingered grippers have been gaining momentum across applications. However, because of their infinite-dimensionality and non-linearities, necessary modeling and control algorithms are difficult to derive. The online modeling and control method proposed in this paper relies on Koopman operator theory for data-driven modeling and online MPC for control. Our proposed method can achieve solid prediction accuracy while training significantly faster compared to related data-driven methods currently in-use. 

Our method can be directly embedded into model-based online control of robotics systems, without requiring explicit knowledge of system dynamics except for their configuration space or workspace. 
Results by testing with two soft 3-fingered robotic grippers attest that our method can grasp a variety of objects of different shape and/or weight without any a-priori knowledge. This work can be beneficial in applications like smart manufacturing and precision agriculture. 


\begin{table}[t]
\vspace{4pt}
    \centering
    \caption{Success rates of grasping different objects }\label{tab:success_rate}
    \vspace{-6pt}
    \begin{tabular}{ c c c c c c c }
    \toprule
     Object & 1&2&3&4&5&6\\
    \midrule
    Gripper & \multicolumn{6}{c}{Symmetric Gripper}\\
    \midrule
     Success Rate & $100\%$ & $100\%$ & $90\%$ & $10\%$ & $0\%$&$0\%$ \\
    \midrule
    \midrule
     Gripper & \multicolumn{6}{c}{Asymmetric Gripper}\\
    \midrule
     Success Rate & $90\%$ & $80\%$ & $100\%$& $90\%$& $70\%$&$0\%$ \\
    \bottomrule
    \end{tabular}
    \vspace{-18pt}
\end{table}

Future work aims to combine the online modeling approach with other model-based controllers (e.g.,~\cite{Mucchiani2022closed}), and develop an approach for generating and adapting online the desired trajectories to grasp a wider range of objects.


\bibliographystyle{IEEEtran}
\bibliography{IEEEabrv,IEEEexample}

\end{document}